\documentclass[journal]{IEEEtran}

\usepackage{graphics} 
\usepackage{epsfig} 
\usepackage{amssymb}  
\usepackage{hyperref}
\usepackage{color}
\usepackage{multirow}
\usepackage[caption=false, font=footnotesize]{subfig}
\usepackage{threeparttable}

\usepackage[cmex10]{amsmath}

\newcommand{\bs}[1]{\boldsymbol{\mathrm{#1}}}

\begin{document}
\title{Reinforcement Learning with Parameterized Manipulation Primitives for Robotic Assembly}

\author{Nghia Vuong$^{1}$, and Quang-Cuong
  Pham$^{1,2}$
  \thanks{$^{1}$Singapore Centre for 3D Printing (SC3DP), School of
    Mechanical and Aerospace Engineering, NTU, Singapore}
  \thanks{$^{2}$Eureka Robotics, Singapore}
}

%


\maketitle

\begin{abstract}
	A common theme in robot assembly is the adoption of Manipulation Primitives as the atomic motion to compose assembly strategy, typically in the form of a state machine or a graph. While this approach has shown great performance and robustness in increasingly complex assembly tasks, the state machine has to be engineered manually in most cases. Such hard-coded strategies will fail to handle unexpected situations that are not considered in the design. To address this issue, we propose to find dynamics sequence of manipulation primitives through Reinforcement Learning. Leveraging parameterized manipulation primitives, the proposed method greatly improves both assembly performance and sample efficiency of Reinforcement Learning compared to a previous work using non-parameterized manipulation primitives. In practice, our method achieves good zero-shot sim-to-real performance on high-precision peg insertion tasks with different geometry, clearance, and material.
\end{abstract}



%
\IEEEpeerreviewmaketitle

\section{Introduction}

A common theme in robot assembly is the adoption of Manipulation Primitives (MP) as the atomic motion to compose assembly strategy, typically in the form of a state machine or a graph \cite{thomasErrortolerantExecution2003}, \cite{finkemeyerExecutingAssembly2005}, \cite{suarez-ruizFrameworkFine2016}. While this approach has shown great performance and robustness in increasingly complex assembly tasks \cite{suarez-ruizCanRobots2018a}, the state machine has to be engineered manually in most cases. Such hard-coded strategies will fail to handle unexpected situations that are not considered in the design. Furthermore, designing and fine-tuning such strategies require considerable engineering expertise and time.

\begin{figure}[!t]
	\centering
	\includegraphics[width=\columnwidth]{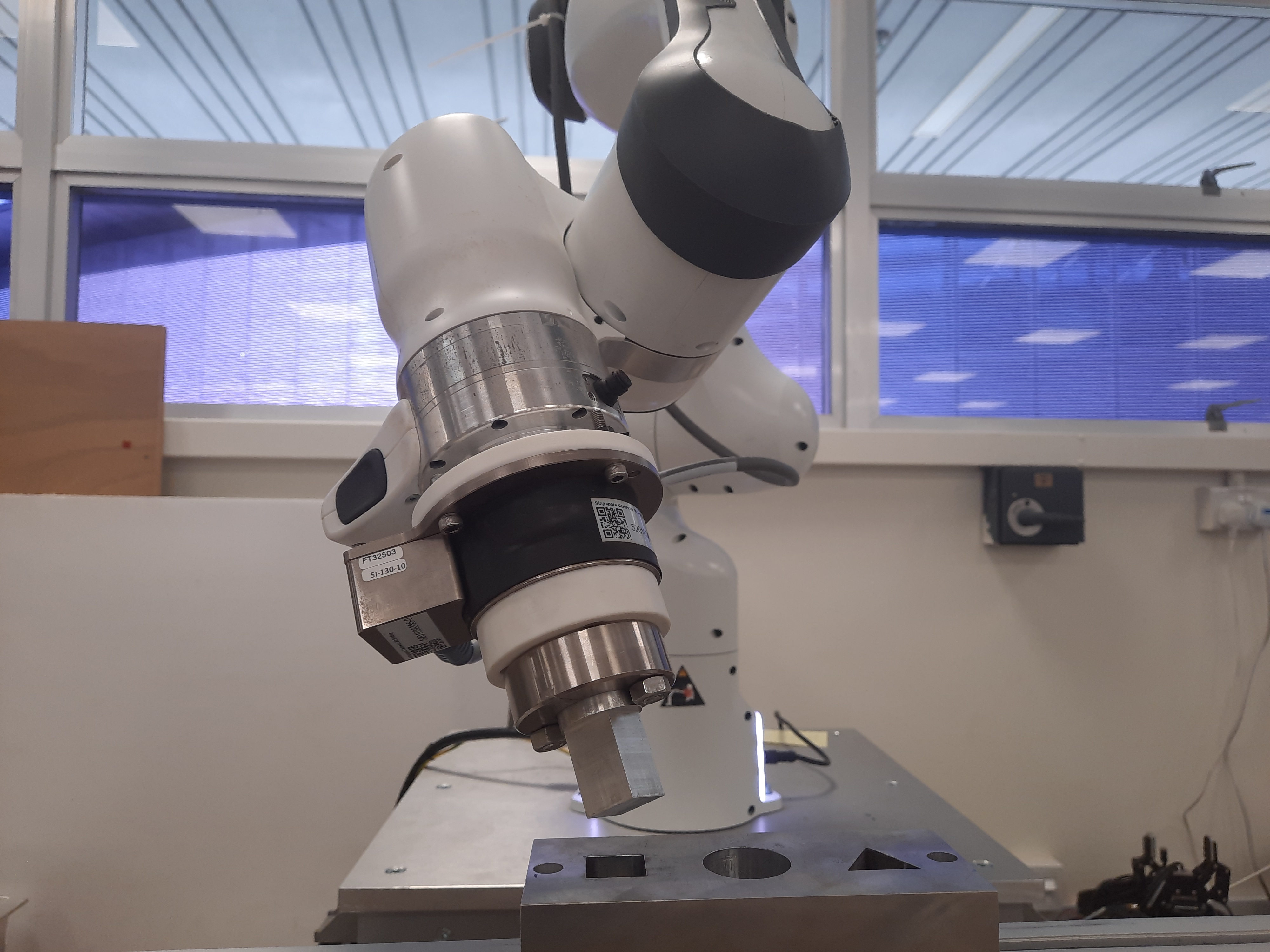}
	\caption{Robotic assembly setup}
	\label{visual-id}
\end{figure}

We are interested in the automatic discovery of assembly strategy, specifically in the form of a dynamics sequence of manipulation primitives. By dynamic sequence, we mean a policy that outputs an MP online from a set of MPs. We propose to learn such policies through Reinforcement Learning (RL), which was initially explored by our group in \cite{vuongLearningSequences2021}\footnote[3]{Here we summarize the main evolutions of this paper with respect to the conference version \cite{vuongLearningSequences2021}\begin{itemize}
  \item The introduction of continuously-parameterized MPs as the action space for RL.
  \item More extensive experimental results showing that parameterized MPs improve RL sample efficiency and task performance.
\end{itemize}}. However, one limitation of this work is the lack of flexibility as the number of MPs in the set is large while only several MPs are useful for a particular task. In this paper we address this fundamental problem by utilizing a set of \emph{continuously-parameterized} MPs. The action space, as a consequence, becomes hybrid: the RL policy needs to output an MP and its parameters (e.g., the speed of movement, the amount of force to terminate an MP). Using parameterized MPs enables reducing the number of MPs from 91 MPs to 13 MPs and expands the range of possible robot motion realized by the set of MPs. The addition of continuous parameters, however, could introduce additional complication for RL algorithm. We show experimentally that parameterized MPs improve both RL sample efficiency and assembly performance. 

Another key advantage of MPs is their additional \emph{semantics}, which make them robust in sim-to-real and against model/environment variations and uncertainties: consider how ``Move down until contact'' is inherently more robust than a sequence of several short ``Move down'' actions. In practice, our method achieves good zero-shot sim-to-real performance on high-precision peg insertion tasks with different geometry, clearance, and material.

The rest of the paper is organized as follows. In Section
\ref{sec-review}, we review works that are related to our proposed
approach. In Section \ref{sec-background}, we introduce the background for the study. In Section \ref{sec-learn-seq}, we detail the procedure for designing a set of MPs and learning assembly strategy with RL. In Section
\ref{sec-exp}, the experimental setup and quantitative
results are presented. Finally, in Section \ref{sec-sum}, we discuss the advantages and limitations of the presented approach, as well as some directions
for future work.

\section{Related Work} \label{sec-review}

\textbf{Manipulation Primitives in robotic assembly.} Manipulation primitive or skill primitive is commonly used as an interface between task specification and control \cite{krogerTaskFrame2004}, \cite{krogerManipulationPrimitives2011} in robot manipulation. Decades-long research have seen wide adoption of MPs in robot assembly \cite{thomasErrortolerantExecution2003}, \cite{finkemeyerExecutingAssembly2005}, \cite{suarez-ruizFrameworkFine2016}. In these works, the assembly strategy is represented as a sequence or a graph composing of MPs. Recently, Johannsmeier et al. \cite{johannsmeierFrameworkRobot2019} propose a method to automatically optimize the parameters for the MPs in a predefined graph to minimize the assembly time through gradient-free optimization methods. It was shown that the graph of MPs with learned parameters outperforms human on performing a sequence of cylindrical pin insertion tasks. A shared limitation of  the above studies is that the sequence of graph of MPs are defined manually, thus lack the ability to generalize to different contexts. Furthermore, since the graph is generated offline, it could not adapt to environmental uncertainties that might occur during execution: one failure of any MPs might lead to the failure of the whole execution. In our method, we do not assume a fixed sequence of MPs; instead, the MP is generated at run time.

\textbf{RL for high precision robotics assembly.} 

Recent studies have explored several methods to improve sample efficiency and performance of RL algorithms in high precision robotics assembly. One such method is smart action space discretization and problem
decomposition. In
\cite{inoueDeepReinforcement2017}, Inoue et al. decompose the task into a search
phase and an insertion phase, and designs different state space, action
space, and reward function for each phase. More specifically, several
meta-actions are designed to form a discrete action space. This
greatly speeds up training and achieves good performance at the same
time. However, one drawback of this method is flexibility, since a few
meta-actions limit the dynamic capabilities of the robot. We address
this problem by using a large set of such meta-actions. Although this
choice compromises the reduced training time, we adopt sim2real to
address this issue and argue that the use of MPs as meta-actions makes
the sim-to-real transfer more efficient. In the same vein, Hamaya et
al. \cite{hamayaLearningRobotic2020a} divide the peg-in-hole task into five
steps with different action spaces and state spaces. This
decomposition greatly speeds up training through dimensional reduction
of action and state spaces. The method, however, assumes a fixed
sequence of steps that might not generalize well to other
tasks.

Another class of methods exploits a model-based or a heuristic policy to improve exploration phase of RL algorithm. One such technique is Residual Reinforcement Learning \cite{johanninkResidualReinforcement2019a}, in which the RL policy is the sum of a hand-designed controller and a parametric policy. Only the latter is learned by RL algorithm, while the former quickly guides the RL agent to the region in state space with high reward. These approaches have been shown to outperform vanilla RL in assembly tasks \cite{johanninkResidualReinforcement2019a}, \cite{schoettlerMetaReinforcementLearning2020a}. Subsequently, several works have studied the use of different types of policies, such as Gaussian Mixture Model \cite{maEfficientInsertion2021}, Dynamic Movement Primitive \cite{davchevResidualLearning2021} in place of the hand-designed controller. 

In \cite{luoReinforcementLearning2019a}, Luo et
al. uses a model-based method, namely iterative Linear-Quadratic-Gaussian (iLQG)
\cite{todorovGeneralizedIterative2005} to find local control policies, then train a neural network that generalizes this controller to adapt to environmental variations, taking into account the force feedback. This method is
fast, being able to find a good control policy in just a few
interactions, but relies on iLQG to find the local controllers that
might not achieve a good performance on complex tasks, e.g. tight
insertion tasks, due to the imposed linear structure on the system
dynamic.

We believe that the idea of using a model-based or heuristic policy can be applied to improve the sample efficiency of our method. For example, one can design a simple sequence of MPs and initialize the control policy to this simple sequence. However, this remains to be explored in future works.

One line of research utilizes sim-to-real techniques. Kaspar et al. \cite{kasparSim2RealTransfer2020a} perform system identification to find several simulation parameters (gravity, joint damping, etc.) to align the simulated with the physical system. However, this approach still require collecting excitation trajectories on the physical system. Domain randomization \cite{tobinDomainRandomization2017}, \cite{pengSimtorealTransfer2018} is another technique that is largely adopted for sim-to-real transfer \cite{schoettlerMetaReinforcementLearning2020a}, \cite{beltran-hernandezVariableCompliance2020}. Domain randomization aims to learn a policy that is robust to task variations, such as position or physical properties of objects. However, domain randomization requires prior knowledge to determine which parameters to randomize and their ranges in order to learn a policy that is robust to sim-to-real gap while also not being too conservative. In contrast, our method relies on the robustness of the MPs for sim-to-real transfer.

\textbf{Reinforcement learning with structured action space}

The choice of action space of RL environment is an important design factor. Common action spaces such as joint space (an action could be desired joint position), task space (an action could be desired end-effector position or robot impedance) have been explored in the literature \cite{martin-martinVariableImpedance2019}, \cite{bogdanovicLearningVariable2020}. To further improve the sample efficiency and performance of RL, recent works have investigated the use of high-level action, such as skills \cite{chitnisEfficientBimanual2020}, modular controllers \cite{sharmaLearningCompose2020}, or motion primitives \cite{vuongLearningSequences2021}, \cite{nasirianyAugmentingReinforcement2021}, \cite{zhangLearningInsertion2021}, \cite{dalalAcceleratingRobotic2021}.

Chitnis et al. \cite{chitnisEfficientBimanual2020} proposes to learn task schema composing of skills, such as grasping, goal reaching, lifting, to solve more complex tasks. However, the task schema is state-independent, which has been shown to struggle with complex manipulation tasks such as nut assembly and peg insertion \cite{nasirianyAugmentingReinforcement2021}.

Sharma et al. \cite{sharmaLearningCompose2020} defines multiple controllers that act on elementary axes of the task frame. The RL agent learns to compose multiple controller in parallel to realize more diverse behaviors. In their formulation, the controller is executed for a fixed number of steps while we consider more general MPs with an explicit condition of termination.

Recently, \cite{dalalAcceleratingRobotic2021} defines action primitive based on a goal state and an error metric. They evaluate their methods extensively in simulation on various benchmarks of robot manipulation tasks. In contrast, we aim to demonstrate our method on the real robot and focus on the high-precision peg insertion tasks.

Most related to our work is \cite{zhangLearningInsertion2021}, where they also propose to learn parameterized motion primitives for assembly tasks. In this work, we consider manipulation primitives with additional semantics with two possible termination conditions. This also provides an additional intrinsic reward to encourage the RL agent to discover consistent strategy. Furthermore, we evaluate our method on the tight peg insertion with square and triangle shape. We observe that these tasks are more challenging than the round peg insertion task.

\section{Background} \label{sec-background}

\subsection{Manipulation primitives} \label{sec-mp}

An MP represents a desired motion of the robot end-effector ($E$) with respect to a task frame ($T$) \cite{vuongLearningSequences2021}. It consists of a desired velocity command $\bs{v}_{\mathrm{des}}$, a desired force command $\bs{f}_{\mathrm{des}}$, and a stopping condition $\lambda$. The desired velocity and force commands are defined as
\begin{align}
	\begin{split}
	\bs{v}_{\mathrm{des}}(t) & := g_v(t, \bs{\Omega}_t;\bs{\theta}_v), \\
	\bs{f}_{\mathrm{des}}(t) & := g_f(t, \bs{\Omega}_t;\bs{\theta}_f), 
	\end{split}
	\label{equation:primitive}
\end{align}
where $g_v$ and $g_f$ are any functions parameterized respectively by
$\bs{\theta}_v$ and $\bs{\theta}_f$, and $\bs{\Omega}_t$ is the vector
of all sensor signals at time $t$. The stopping condition is
defined as
$\lambda : (t,\bs{\Omega}_t, \bs{\theta}_s) \mapsto \{\verb+SUCCESS+, \verb+FAILURE+,
\verb+CONTINUE+\}$. An MP terminates when \(\lambda\) returns either \verb+SUCCESS+ or \verb+FAILURE+. An MP is fully defined by the functions \(g_v\) \(g_f\), \(\lambda\), and their parameters \(\bs{\theta}=[\bs{\theta}_v, \bs{\theta}_f, \bs{\theta}_s]\)

Executing an MP requires controlling of desired end-effector velocity/position and force, which can be realized by the hybrid motion/force control scheme. In this scheme, the desired velocity and desired force is achieved by a position control loop and a force control loop, respectively. In this paper, we use the inverse dynamics control in the operational space approach \cite{khatibUnifiedApproach1987} for the position control loop and a simple feedforward force controller for the force control loop.

\subsection{Reinforcement learning} \label{sec-rl}

We consider here the discounted episodic RL problem. In this setting,
the problem is described as a Markov Decision Process (MDP)
\cite{suttonReinforcementLearning2018a}. At each time step $t$, the agent
observes current state $\bs{s}_t \in \mathcal{S}$,
executes an action $\bs{a}_t \in \mathcal{A}$, and receives an immediate
reward $r_t$. The environment evolves through the state transition
probability $p(\bs{s}_{t+1} | \bs{s}_t, \bs{a}_t)$. The goal in RL is to learn a
policy $\pi (\bs{a}_t|\bs{s}_t)$ that maximizes the expected discounted return
$R = \sum_{t=1}^{T} \gamma^t r_t$, where $\gamma$ is the discount
factor.

\subsection{Reinforcement learning with parameterized action space} \label{sec-pamdp}

A parameterized action space consists of a finite set \(\mathcal{A}_d=\{a_1,a_2\dots a_n\}\) with cardinality \(n\) and \(n\) sets \(\mathcal{X}_i, \; i=\overline{1,n}\). An action is a tuple \((a_i, x)\), where \(a_i\in \mathcal{A}_d\) and \(x\in \mathcal{X}_i\). Problems with parameterized action space can be formally defined as parameterized action MDP (PAMDP) \cite{massonReinforcementLearning2016}. In a PAMDP, the RL agent makes a decision in two steps: it first s \(a_i\) and then choose \(x\). One can think of the variable \(x\) as some parameters of the discrete action \(a_i\). For example, consider the manipulation primitive defined in \ref{sec-mp}, \(x\) can be the parameters of an MP. Finally, the policy can be written as

\begin{equation}
  \pi(a_i,x|s)=\pi^d(a_i|s_t)\pi^c_i(x|s_t)
  \label{eq-policy}
\end{equation}
where \(\pi^d(a_i|s_t)\) is denoted as the discrete-action policy and \(\pi^c_i(x|s_t)\) is denoted as the action-parameter policy

\section{Learning sequences of manipulation primitives for peg insertion tasks} \label{sec-learn-seq}

Manipulation primitives have long been used as the atomic actions to compose manipulation skills \cite{finkemeyerExecutingAssembly2005}, \cite{krogerManipulationPrimitives2011}, \cite{suarez-ruizFrameworkFine2016}. Despite using slightly different definitions of MP, the main idea is to provide a simple and unified interface to task specification and robot programming. Based on this idea, we propose a method to automatically find the sequences of MPs and their parameters for a class of high-precision peg insertion tasks. Note that the methodology detailed in this section could be applied for tasks other than peg insertion.

We make the following assumptions regarding the peg insertion task

\begin{figure}[!t]
	\centering
	\subfloat[Free-space, translate until contact]{
		\label{fig-primitive-a}
		\includegraphics[width=0.4\columnwidth]{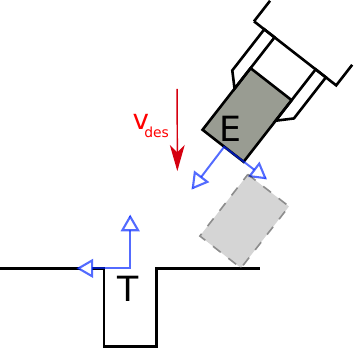}}
	\hspace{0.1\columnwidth}
	\subfloat[Free-space, translate a predefined distance]{
		\label{fig-primitive-b}
		\includegraphics[width=0.4\columnwidth]{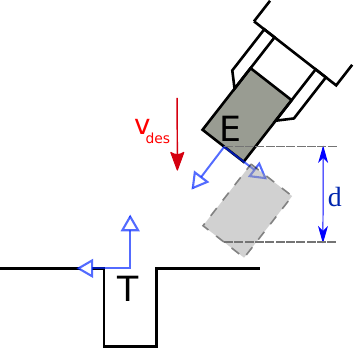}}\\
	\subfloat[In-contact, rotate until next contact]{
		\label{fig-primitive-c}
		\includegraphics[width=0.27\columnwidth]{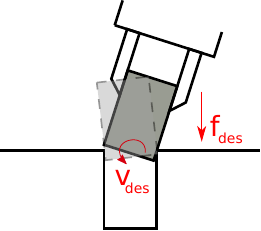}}
	\hspace{0.1\columnwidth}
	\subfloat[In-contact, insert]{
		\label{fig-primitive-d}
		\includegraphics[width=0.27\columnwidth]{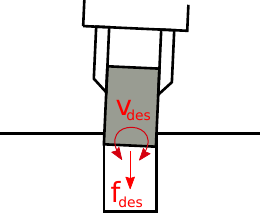}}
	\caption{Examples of Manipulation Primitives for insertion task}
	\label{fig-primitive}
\end{figure}

\begin{itemize}
  \item The peg is firmly grasped or rigidly attached to the end-effector
  \item The hole position (defined as the point T in Fig.~\ref{fig-primitive-a}) and axis of insertion can be estimated with position error less than 1\,mm and orientation error less than 1\,degree. This can be achieved by, for example, using visual servoing technique \cite{yuSiameseConvolutional2019}.
  \item The task frame is chosen such that its origin coincides with the estimated hole position and the direction of insertion is along the -z axis. This assumption simplifies the design of MPs without the loss of generality.
\end{itemize}

We consider two families of MPs, free-space MPs and in-contact MPs:
\begin{itemize}
	\item Free-space MPs are to be executed when the robot is not in
	      contact with the environment, i.e., when all external forces/torques
	      are zero. MPs in this family are then associated with zero desired
	      force/torque command.
	\item In-contact MPs are to be executed when the robot is in contact
	      with the environment, i.e. when some of the external force/torque
	      components are non-zero. In addition to other objectives, MPs in
	      this family have some of the components of their desired
	      force/torque command to be non-zero in order to maintain the same
	      contact state during the execution. Given the assumption that the direction of insertion is along the -z axis, the desired force/torque command has the form \(\bs{f}_{des}=[0, 0, -f_d, 0, 0, 0]\)
\end{itemize}

Each family of MPs is subdivided into three types: (i) move until
(next) contact, (ii) move a predefined amount, (iii) insert. Figure
\ref{fig-primitive} illustrates some examples of MPs, which are
further detailed below.

\textbf{Move until contact.} Translate the end-effector
along a direction \(\bs{u}\), or rotate the end-effector about a direction \(\bs{u}\), until contact is detected. The stopping condition is \texttt{SUCCESS} if \(\bs{f}_{\mathrm{ext}}^T\bs{u}>f_\mathrm{thr}\), \texttt{FAILURE} if the execution time is larger than \(T\) secs. The parameters of this MP is \((\bs{u},v,f_\mathrm{thr},T)\), where \(v\) denotes motion speed. The parameters include the desired force \(f_d\) if the MP is an in-contact MP.

\textbf{Move fixed distance.} Translate the
end-effector along a direction \(\bs{u}\) over a predefined distance~$d$, or
rotate about a direction \(\bs{u}\) over a predefined angle~$d$. The stopping condition is \texttt{SUCCESS} if robot reaches the desired distance/angle, \texttt{FAILURE} if interaction force/torque exceeds \(f_\mathrm{max}\). The parameters of this MP is \((\bs{u},d,v,f_\mathrm{max})\) where \(v\) denotes the motion speed. The parameters include the desired force \(f_d\) if the MP is an in-contact MP.

\textbf{Insert.} Track a non-zero force in the direction
of insertion and regulate the external torques to zero in all the
directions. To regulate torques to zero, we set desired velocity as \(\bs{v}_{\mathrm{des}}(t) = -K_d\bs{f}_{\mathrm{ext}}\) where \(K_d\) is a diagonal matrix and its diagonal is \(diag(K_d)=[0,0,0,k,k,k],\;k\geq0\). The stopping condition is \texttt{SUCCESS} if the end-effector reach a goal position within 2\,mm, \texttt{FAILURE} if the execution time is larger than \(T\) secs. The parameters of this MP is \((f_d,k,T)\)

Instead of manually design a sequence of MPs and tune their parameters for a particular tasks, we propose to automatically find the sequence of MPs and their parameters through RL. For this purpose, we first design a set of MPs and define bounds (or values) for all parameters. An example is shown in table~\ref{tab-mp-hybrid}, which consists of 13 MPs parameterized by 18 parameters. The axis of movement \(\bs{u}\) is chosen from the set of elementary axes \(x,y,z\) of the task frame. Note that other design choices are possible. For example, one can consider the axis of movement as a continuous parameter as in \cite{zhangLearningInsertion2021}.

\begin{table*}[!ht]
	\vspace*{5pt}
	\caption[caption]{The set of 13 Manipulation Primitives used in experiment. Ranges correspond to learnable parameters}
	\label{tab-mp-hybrid}
	\begin{center}
		\begin{threeparttable}
			\begin{tabular}{|c|c|c|c|c|c|}
				\hline
				Family                      & Type                                                      & Axis                                   & Parameters       & Value/Range             & N                   \\
				\hline
				\multirow{8}{*}{Free space} & \multirow{3}{*}{Translate until contact (T$^\textrm{c}$)} & \multirow{3}{*}{$-z$}                  & $v$              & 10\,mm/s           & \multirow{3}{*}{1}  \\
				                            &                                                           &                                        & $f_\mathrm{thr}$ & 5\,N               &                     \\
                                    &                                                           &                                        & $T$ & 2\,s               &                     \\

				\cline{2-6}
				& \multirow{3}{*}{Translate (T)} & \multirow{3}{*}{$x,y$} &
				$v$ & 10\,mm/s & \multirow{3}{*}{2} \\
				                            &                                                           &                                        & $f_\mathrm{thr}$ & 20\,N              &                     \\
				                            &                                                           &                                        & $d$              & [-10, 10]\,mm         &                     \\
				\cline{2-6}
				& \multirow{3}{*}{Rotate (R)} & \multirow{3}{*}{$ x,
				y$} &
				$v$
				& 0.1\,rad/s & \multirow{3}{*}{2} \\
				                            &                                                           &                                        & $f_\mathrm{thr}$ & 2\,Nm              &                     \\
				                            &                                                           &                                        & $d$              & [-0.1, 0.1]\,rad        &                     \\
				\hline
				\multirow{17}{*}{In contact} & \multirow{4}{*}{Translate
				until next
				contact (T$^\textrm{c}$)}& \multirow{4}{*}{$ x,  y$} &  $v$ & [-10, 10]\,mm/s & \multirow{4}{*}{2}\\
				
				                            &                                                           &                                        & $f_\mathrm{thr}$ & [5, 12]\,N         &                     \\
				                            &                                                           &                                        & $f_d$            & -8\,N              &                     \\
                                    &                                                           &                                        & $T$ & [0.1, 2]\,s               &                     \\
				\cline{2-6}
				
				
				                            & \multirow{4}{*}{Translate (T)}                            & \multirow{4}{*}{$ x,  y$}        & $v$              & [5, 10]\,mm/s           & \multirow{4}{*}{2}  \\
				
				                            &                                                           &                                        & $f_\mathrm{thr}$ & 20\,N              &                     \\
				                            &                                                           &                                        & $d$              & [-10, 10]\,mm         &                     \\
				                            &                                                           &                                        & $f_d$            & $-8$\,N            &                     \\
				\cline{2-6}
				
				                            & \multirow{4}{*}{Rotate (R)}                               & \multirow{4}{*}{$ x,  y,  z$} & $v$              & 0.1\,rad/s         & \multirow{4}{*}{3} \\
				
				                            &                                                           &                                        & $f_\mathrm{thr}$ & 2\,Nm              &                     \\
				                            &                                                           &                                        & $d$              & [-0.1, 0.1]\,rad        &                     \\
				                            &                                                           &                                        & $f_d$            & $-8\,$N            &                     \\
				\cline{2-6}
				                            & \multirow{5}{*}{Insert (I)}                               & \multirow{4}{*}{$-z$}                  & $\epsilon$       & 2\,mm                & \multirow{5}{*}{1}  \\
				                            &                                                           &                                        & $k$         & [0.01, 0.2]        &                     \\
				                            &                                                           &                                        & $f_d$            & [6, 15]\,N   &                     \\
                                    &                                                           &                                        & $T$ & [0.1, 2]\,s               &                     \\
				\hline
			\end{tabular}
		\end{threeparttable}
	\end{center}
	
	\vspace*{-10pt}
\end{table*}

Given a set of MPs, the problem of finding a sequence of MPs can be naturally formulated as a PAMDP where an action is \((a_i,x)\); the discrete action \(a_i\) is an MP and \(x\) is its parameter. The state is defined by
$\bs{s}_t = [\bs{p}_t, \bs{f}_{ext, t}]$, where $\bs{p}_t$
is the pose of the peg relative to the hole (position and orientation of frame E with respect to frame T in Fig.~\ref{fig-primitive-a});
$\bs{f}_{ext, t}$ is the external force and torque acting on the
end-effector. Note that at a particular state, not all MPs are feasible. For example, in-contact MPs shouldn't be executed while the robot is in free space. Denote $\mathcal{A}_\mathrm{free}$ the set of free-space MPs and $\mathcal{A}_\mathrm{con}$ the set of in-contact MPs, the set of feasible actions at each state is either $\mathcal{A}_\mathrm{free}$ if $f_\mathrm{ext,z}=\bs{0}$, or $\mathcal{A}_\mathrm{con}$ if $f_\mathrm{ext,z}\neq\bs{0}$.

The reward function is defined as

\begin{eqnarray}
	r_t:=  c_1\left(e^{\frac{-||\bs{p}_{t+1} - \bs{p}_{goal}||_2^2}{k_1}} -1\right)+ c_2s(\bs{a}_t)
	\label{eq-rew}
\end{eqnarray}
where \(c_1, c_2, k_1\) are positive constants. The first term rewards the RL agent for moving closer to the goal, while second term $s(\bs{a}_t)=0$ if
a \verb+SUCCESS+ status is returned, $s(\bs{a}_t)=-1$ if a \verb+FAILURE+
status is returned. This term can be thought as an intrinsic reward to maintain the semantics of the MPs.

The episode terminates when the number of MPs exceed 15 or the task is successful. In the latter case, the RL agent receives an additional termination reward of 5. The task is considered successful if the end-effector reach the goal position within 2\,mm.

\begin{figure}[!t]
	\centering
  \includegraphics[width=\columnwidth]{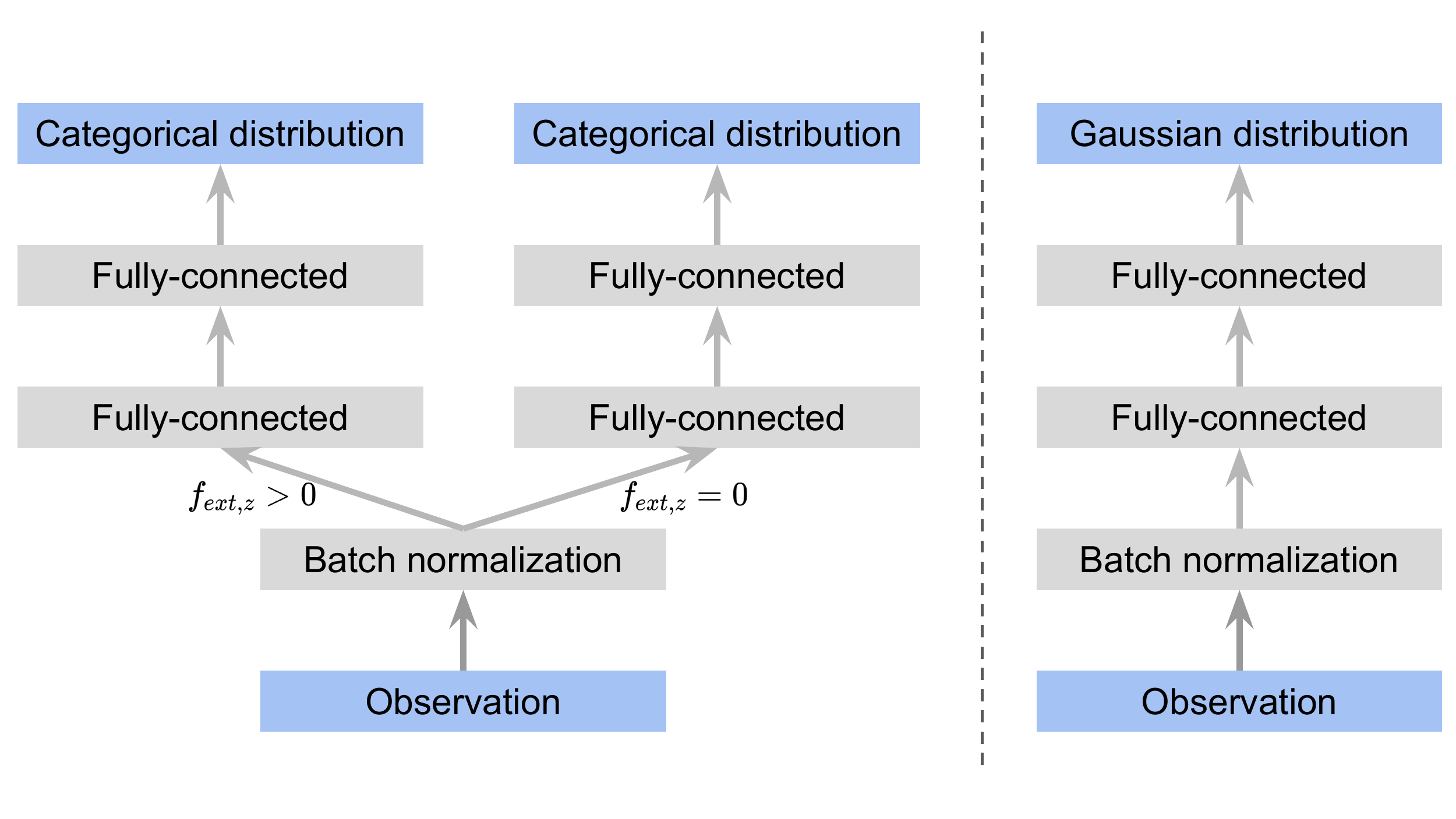}
  \caption{Discrete-action policy network (left of dash line) and action-parameter policy network (right). The networks share a layer of batch normalization. The discrete-action policy contains two separate networks, one for each action subspace.}
	\label{fig-policy}
\end{figure}

\textbf{Policy architecture} As mentioned in \ref{sec-pamdp}, the policy in a PAMDP is composed of a discrete-action policy and an action-parameter policy. The discrete-action policy outputs the logits that define a categorical distribution over all MPs, while the action-parameter policy outputs the mean of a Gaussian distribution over the parameters of the current MP. To realize the state-dependent action space, the discrete-action policy is represented by two Neural Networks (NNs) which output the logits for the in-contact and free-space MPs, respectively. Given an input, the policy performs forward propagation only on one network depending on the measured external force. The action-parameter policy consists of multiple heads, one for each MP. The architecture of NNs are shown in Fig.~\ref{fig-policy}.

The policy is trained with PPO \cite{schulmanProximalPolicy2017}. In addition to the policy network, PPO requires a value network to approximate the value function. The value network has similar network architecture to the action-parameter networks and has independent weights.

\section{EXPERIMENTS} \label{sec-exp}

We conduct experiments to (1) validate the proposed method in simulation on various peg insertion with different shape (2) evaluate the sim-to-real performance of the learned policy.

\subsection{Experimental setups}

\textbf{Task description.} We evaluate the proposed approach on high-precision pin insertion tasks with different shapes and different clearances. More details are shown in Table \ref{tab-dim}.

\begin{table}[!t]
	\vspace*{5pt}
	\caption[caption]{Dimensions and material of pegs and holes. Size is diameter for round profile and side length for square and triangle one}
	\label{tab-dim}
	\begin{center}
		\begin{tabular}{|c|c|c|c|c|}
			\hline
			Profile  & Hole size (mm) & Peg size (mm) & Material \\
			\hline
			Round    & 30.03 & 29.96   & Aluminum  \\
			\hline
			Round    & 30.03 & 29.9   & Aluminum  \\
			\hline
			Square   & 19.98 & 19.96    & Aluminum      \\
			\hline
			Square   & 19.98 & 19.72   & Plastic      \\
			\hline			
			Triangle & 25 & 24.9  & Aluminum     \\
			\hline
			Triangle & 25 & 24.2  & Plastic     \\
			\hline
      
		\end{tabular}
	\end{center}
\end{table}

\textbf{Robot system setup.} A 7-DOF Franka Emika Panda robot is
used in our experiment. We additionally attach a Gamma IP60 force torque sensor to the flange of the robot to measure
the external force and torque acting on the end-effector. The measurement from FT sensor is needed to implement the insert primitive, as we observe that the external torque
estimation provided by \verb+libfranka+ is not precise enough to perform this
motion. Controller is implemented based on Robot Operating System and runs at 1000Hz. Execution of manipulation primitives is implemented based on ROS service-client framework.

\textbf{Simulation environment} We use the Mujoco physics engine
\cite{todorovMuJoCoPhysics2012} and adapt an open-source Panda robot model \footnote{available online at \href{https://github.com/vikashplus/franka_sim}{franka\_sim}}. The controller is simulated with a control frequency of 500Hz, similar to the simulation step. We use a lower frequency in simulation than that in real world to increase computational speed, while still maintaining a stable simulation. Three simulation environments are created for each of the peg shape shown in Fig.~\ref{tab-dim}. The \emph{perception error} is simulated by adding positional and rotational noise to the actual hole pose, this ``estimated'' hole pose is then used to compute the task frame for manipulation primitives, observation and reward for RL.

\textbf{RL implementation details.} We use \texttt{gym} \cite{brockmanOpenAIGym2016} to design the RL environment and \texttt{garage} \cite{garage}, an RL framework based on PyTorch for the implementation of PPO algorithm. For the policy architecture, each fully-connected layer has 128 nodes in the discrete-action policy network and 24 nodes in the action-parameter policy networks. The hyperparameters for PPO are listed as follows: clip ratio is 0.1, minibatch size is 64, discount factor is 0.99, learning rate is 0.0005, and policy entropy coefficient is 0.001, 2048 samples are collected for each policy update.

At the start of each episode, the robot's end-effector is reset to a position 10\,mm along the z axis of the task frame. During training, the perception error is varied at the beginning of each episode by sampling the positional uncertainty in \([-1,1]\)\,mm uniformly for all axes. To generate random rotational uncertainty, we sample a random unit vector and a random angle in \([-1, 1]\)\,degree uniformly, which together define an axis-angle representation of the uncertainty rotation.

\textbf{Baselines} We compare our method with three baselines. The first baseline learns in a purely continuous action space. Specifically, the control policy outputs the desired end-effector displacement at a rate of 40\,Hz. Since the OSC runs at a higher frequency, the input to the OSC is interpolated between 0 and the desired end-effector displacement during one policy step. We refer to the first baseline \texttt{ee-pose}. The second baseline learns in a purely discrete action space. The RL agent decides at each step which MP to execute on the robot, while its parameters are pre-defined. We manually pick two values within the range of parameters shown in Fig.~\ref{tab-mp-hybrid} to form a set of 81 MPs. We refer to this baseline \texttt{discrete} and our method \texttt{hybrid}.

The third baseline is based on \cite{johannsmeierFrameworkRobot2019}, where a fix sequence of MPs is manually defined and MPs parameters are optimized through gradient-free optimization techniques. Following \cite{johannsmeierFrameworkRobot2019}, the strategy has five steps (1) approach, (2) contact, (3) fit, (4) align, (5) insertion except for some slight differences. We use OSC as the low-level controller similar to our method and other baselines and use our insert MP in step 5 instead of sinusoidal motion. These modifications reduce the number of parameters included for optimization to four. The parameters are optimized by the Covariance Matrix Adaptation Evolutionary Strategy (CMA-ES) algorithm \cite{hansenCompletelyDerandomized2001}, which shows the best performance in \cite{johannsmeierFrameworkRobot2019} among the considered algorithm. We refer to this baseline \texttt{fix-seq}.

\subsection{Simulation result}

\begin{figure}[!t]
	\vspace*{5pt}
	\centering
  \subfloat[Round]{\includegraphics[scale=0.25]{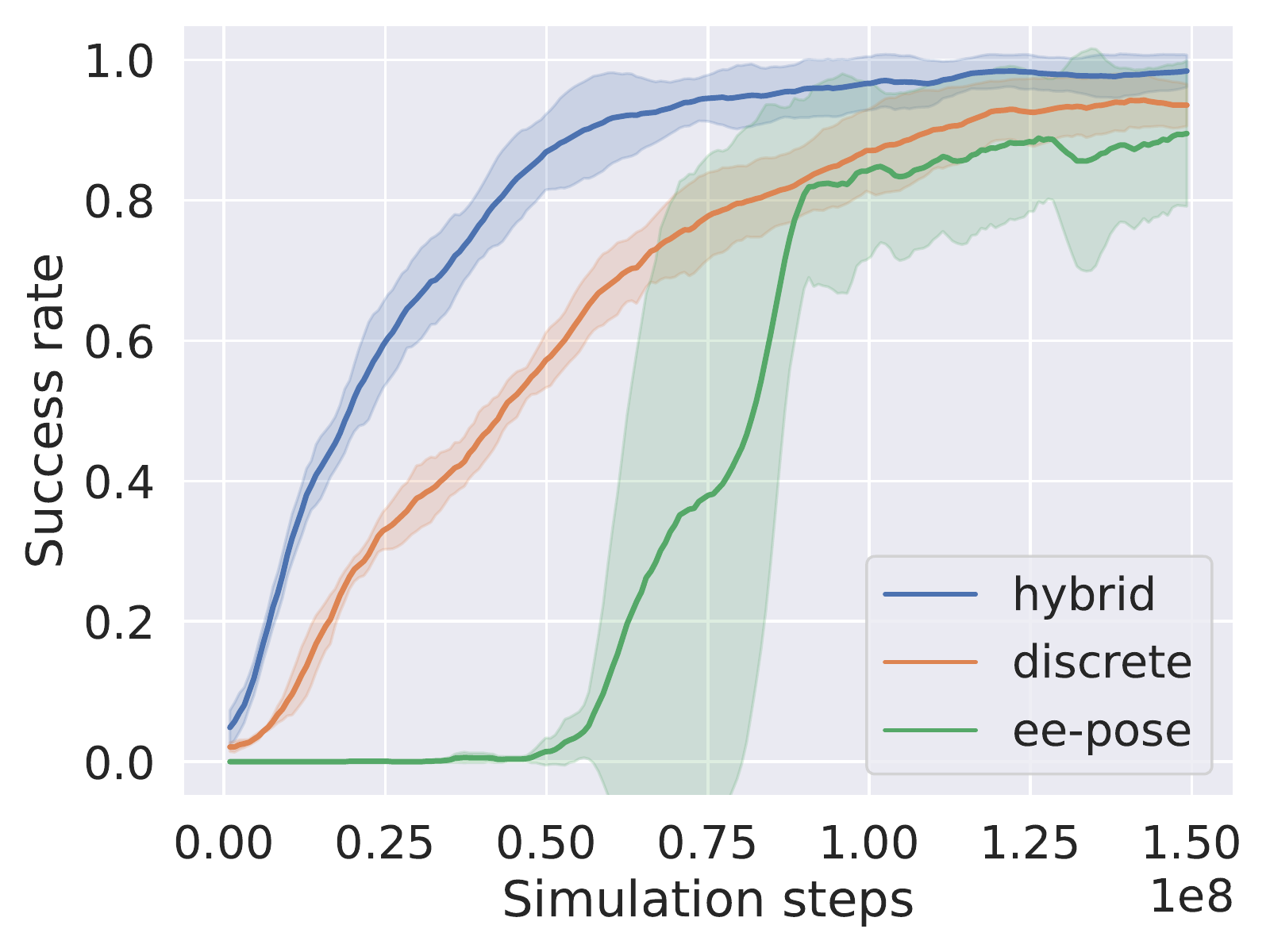}} \hspace{0.0\columnwidth}
  \subfloat[Square]{\includegraphics[scale=0.25]{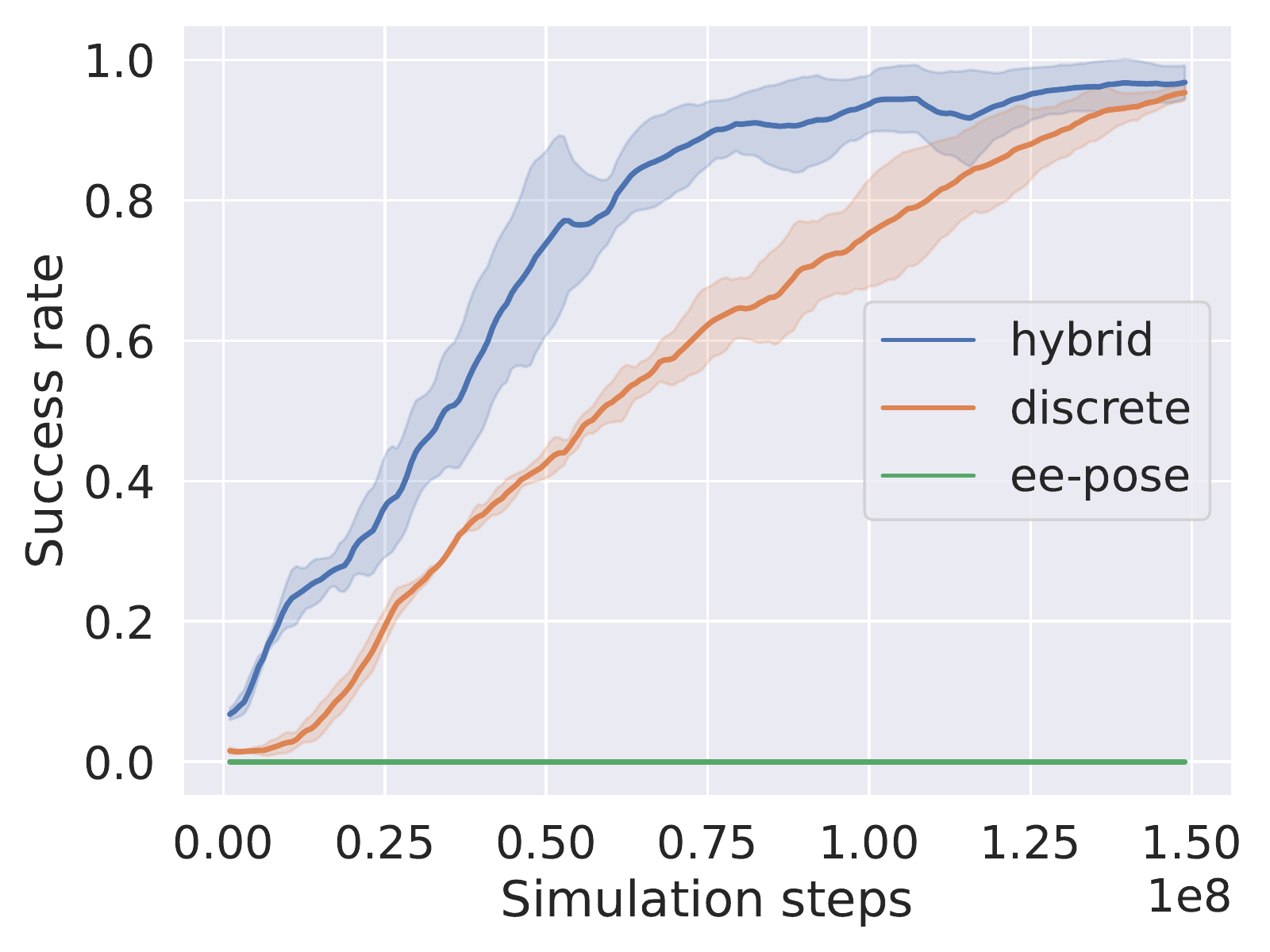}}\\
  \subfloat[Triangle]{\includegraphics[scale=0.25]{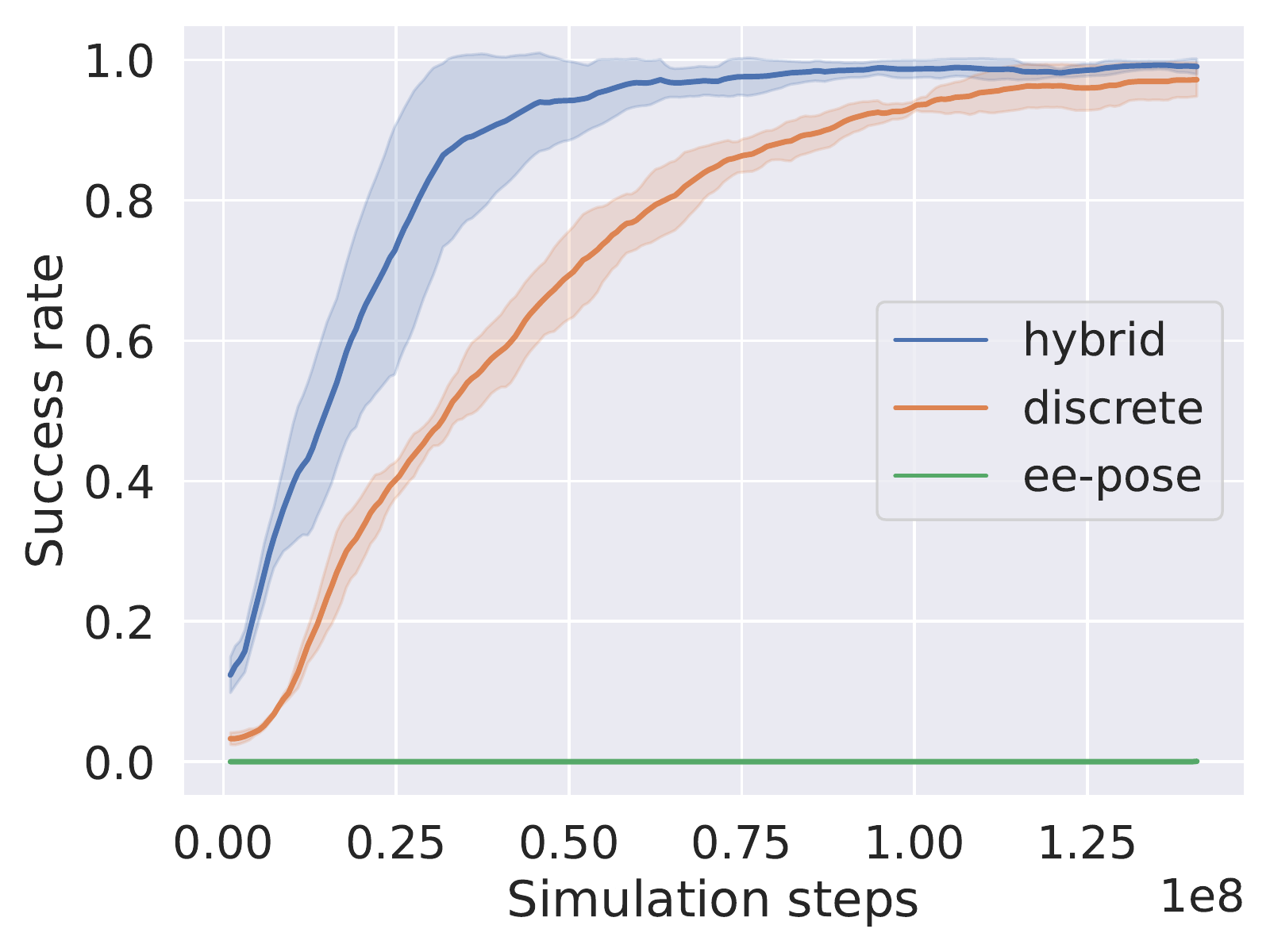}}
  \caption{Training curve showing success rate over cumulative simulation steps. One simulation step corresponding to 2\,ms. The solid curve depicts the mean and the shaded region depicts the standard deviation over three different random seeds.}
	\label{fig-training}
\end{figure}

We train three instances of each method (exclude \texttt{fix-seq}) with three different seeds. The training curves are shown in Fig~\ref{fig-training}. In all three tasks, our method is much more sample efficient and achieve higher final success rate than the other baselines. The number of samples required to reaches 60\% success rate for \texttt{hybrid} is about 2 times less than that of \texttt{discrete}. We also found that \texttt{ee-pose} has trouble in exploration at start of learning. This could be explained by the fact that in high-precision assembly task, the region in state space that leads to the goal is very small. Furthermore, during insertion phase, random exploration noise also causes large interaction force between peg and hole along x and y axes, which prevents the peg to move along the z axis.

\begin{figure}[!t]
	\vspace*{5pt}
	\centering
  \subfloat[Success rate]{\includegraphics[scale=0.3]{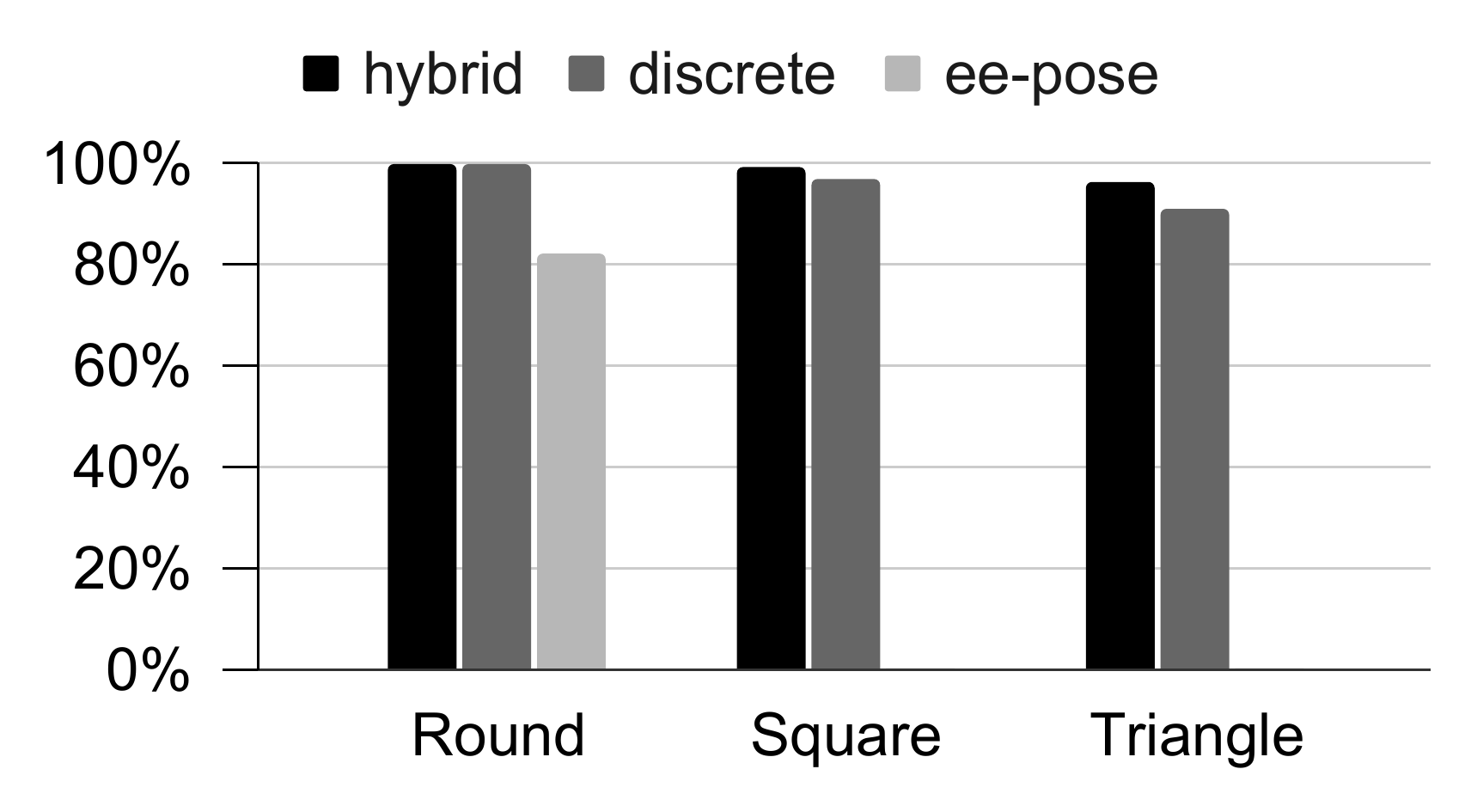}} \hspace{0.0\columnwidth}
  \subfloat[Execution time (s)]{\includegraphics[scale=0.3]{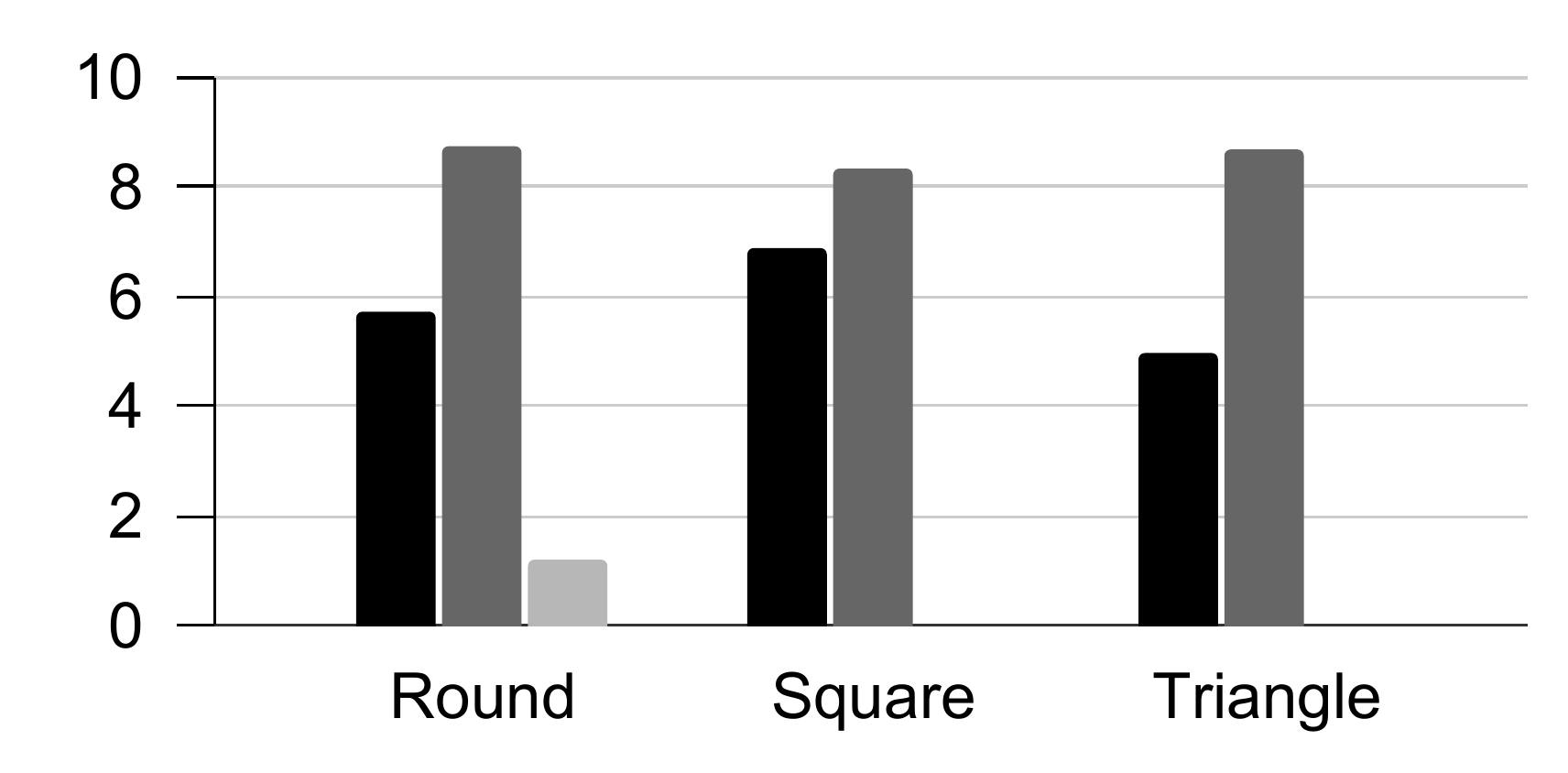}}
  \caption{Quantitative evaluation in simulation. The final success rate and execution time is averaged over 100 trials.}
	\label{fig-eval-sim}
\end{figure}

After training, we obtain the quantitative performance of the trained policies by executing each policy on the corresponding tasks for 100 trials. The average success rate and average execution time is shown in Fig.~\ref{fig-eval-sim}. Overall, \texttt{hybrid} achieves the best performance in terms of success rate, which is consistent with the training curve. Our method also achieves better execution time than \texttt{discrete} as the manually chosen parameters in \texttt{discrete} could be suboptimal. Another observation is that the average execution time for \texttt{ee-pose} is only 1.23\,s, much less than \texttt{hybrid} and \texttt{ee-pose}. There are two reasons for this result. First, the strategy learned by \texttt{hybrid} and \texttt{discrete} has a clear ``search'' phase whose purpose is to align the peg with the hole. The time required for this search phase depends on the environmental uncertainty (the hole pose in this case) causing larger execution time. Second, in \texttt{ee-pose}, robot motion is not constrained as in the other two baselines, i.e. the robot is free to move in any direction, thus the method can explore more diverse behavior. This is a limitation of our method which could be addressed by adding a dummy primitive that moves the robot end-effector to a desired position \cite{dalalAcceleratingRobotic2021}.

\subsection{Sim2real policy transfer on physical robot}
\begin{figure}[!t]
	\vspace*{5pt}
	\centering
  \subfloat[Success rate]{\includegraphics[scale=0.3]{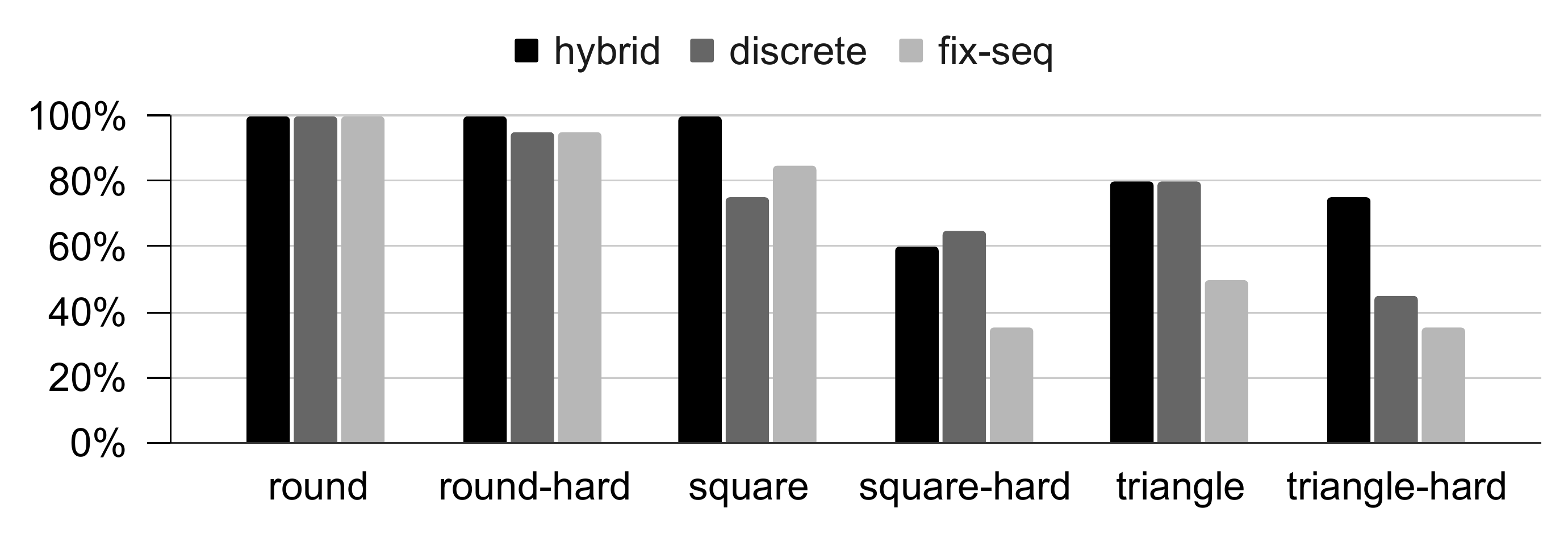}} \hspace{0.0\columnwidth}
  \subfloat[Execution time]{\includegraphics[scale=0.3]{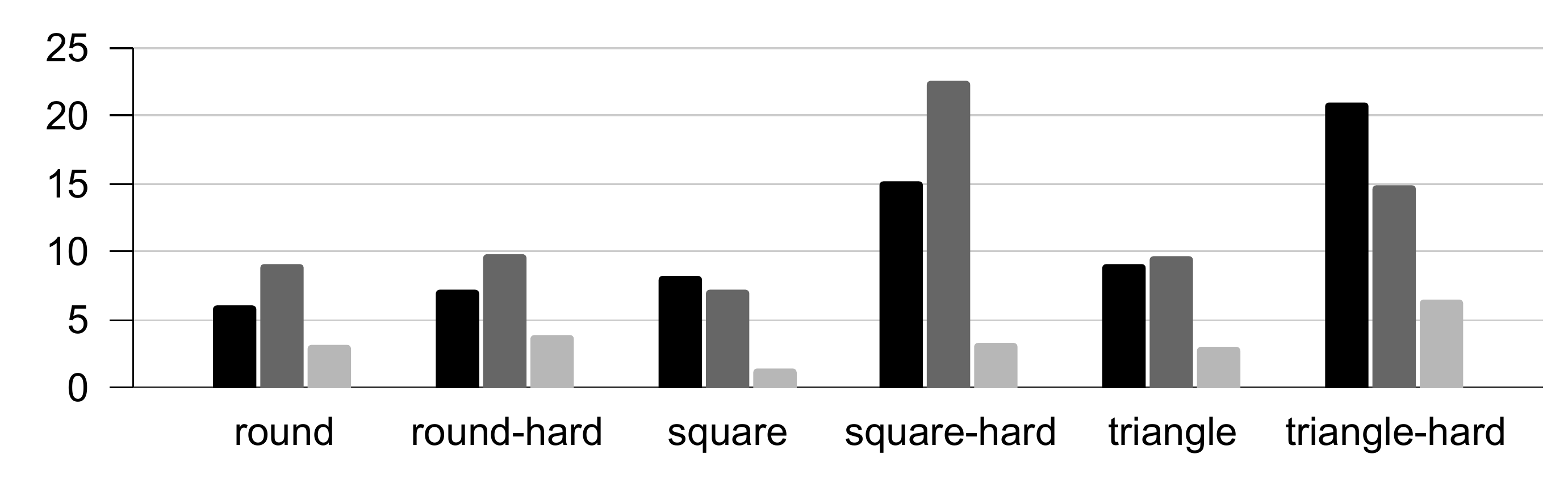}}\\
  \caption{Quantitative evaluation on the physical robot across six different pin insertion tasks. The "hard" suffix denotes tasks with smaller clearance. The final success rate/execution time is averaged over 20 trials}
	\label{fig-eval-real}
\end{figure}

We evaluate the trained policies in simulation directly on six tasks corresponding to six peg-hole pairs shown in Table~\ref{tab-dim} without any further fine-tuning. In this experiment, we also report result for \texttt{fix-seq} baseline. For each task, an optimization is carried out to optimize the parameters of MPs. We then execute the final sequence for 20 trials and report the average success rate and execution time together with other methods. The results are shown in Fig.~\ref{fig-eval-real}. We omit the result for \texttt{ee-pose} because it always fails due to large contact force.

In general, we observe the same pattern as in the simulation: \texttt{hybrid} achieves higher success rate and shorter execution time than \texttt{discrete}. An exception is for the triangle-hard task, \texttt{hybrid} has longer execution time than \texttt{discrete}, but the success rate almost doubles that of \texttt{discrete}. The \texttt{fix-seq} baseline achieves the best performance in terms of execution time. The reason is that the fix sequence always execute four MPs per trial, while RL policies learned by other methods attempt a longer sequence to search for the hole.

We observe two common failure modes during evaluation of the policy learned by our method. First the policy repeatedly choose the ``move until contact'' MP even after the peg has already made contact, causing the robot to be locked in the current contact state. The second failure mode is when the policy fails to fit and align with the hole. We hypothesize that this is because the policy makes decision based solely on the relative pose between peg and hole, which is amenable to sim2real gap in contact modeling (i.e. overlap between two bodies in contact) and difference in object's size. We believe that the first problem could be addressed by incorporating contact force to RL observation, as contact force is more natural signal to infer contact state than pose. However, this remains to be investigated in the future.

\section{Conclusion} \label{sec-sum}
In this paper, we have proposed a method to find dynamics sequence of manipulation primitives through Reinforcement Learning. Leveraging parameterized manipulation primitives, the proposed method was shown to greatly improve both assembly performance and sample efficiency of Reinforcement Learning. The experimental results showed that policies learned purely in simulation were able to consistently solve peg insertion tasks with different geometry and very small clearance.

A limitation of the approach is that the policy depends only on the kinematics information, while interaction force is only used to separate the free and in-contact action subspace. We hypothesize that this is the main causes for sim-to-real transfer failure. Integration of tactile information is thus an interesting future works.

In tasks that have complex dynamics, where instability is a key
consideration, the choice of the low-level control law in each MP
define the upper limit of the overall system performance. Hence,
incorporating advanced robust control laws~\cite{phamConvexController2020} is a
promising direction.



\section*{Acknowledgment}
This research was supported by the National Research Foundation, Prime
Minister’s Office, Singapore under its Medium Sized Centre funding
scheme, Singapore Centre for 3D Printing, CES\_SDC Pte Ltd, and Chip
Eng Seng Corporation Ltd.

\ifCLASSOPTIONcaptionsoff
  \newpage
\fi



\bibliographystyle{IEEEtran}
\bibliography{ref}

\end{document}